\documentclass[preprint,12pt,nopreprintline]{elsarticle}

\usepackage{amssymb}
\usepackage{subcaption}

\usepackage{amsmath}
\usepackage{multirow}

\begin{document}

\begin{frontmatter}



\title{Enhancing Active Learning Performance with Initial Data Selection Based on Foundation Models}



\author{Mohammad Mirzaee, Mehrshad Fallah Estalkhzir, Mohammad Reza Mohammadi} 
\ead{mohammad_mirzaee@comp.iust.ac.ir, fallah_mehrshad82@comp.iust.ac.ir, mrmohammadi@iust.ac.ir}
\affiliation{organization={Iran university of science and technology },
            city={Tehran},
            country={Islamic Republic of Iran}}

\begin{abstract}
In active object detection, selecting the initial labeled pool is essential, especially in complex datasets with expensive labeling and low resource budget. This study presents an active learning framework designed to optimize the initial training cycle of active learning for the object detection task. Our approach diverges from traditional methods by utilizing feature-based clustering from the YOLO-World model and refining images based on objectness scores rather than relying on uncertainty sampling in the initial cycle. This approach focuses on finding samples with various features and emphasizes training in areas where the foundation model is most confident. Doing so helps the model better understand different types of objects. For subsequent cycles, we employ a consistency-based sampling technique to maintain the robustness of the training process. To evaluate our method, we use two complex datasets called Aquarium and NWPU VHR-10, which are well known for their diverse object sizes and challenging conditions. In the NWPU dataset, our method increases mAP50 by an average of 6.6\% in image-based evaluation. Moreover, the image refinement method shows an improvement of approximately 11\% in object-based evaluation. The results show improvements in model performance and labeling time, demonstrating the advantages of using foundation models in early cycles of active learning. An implementation of the algorithm is available at :https://github.com/Mhrshd-fllh/ALBoost-FM
\end{abstract}

\begin{graphicalabstract}
\includegraphics[scale=0.32]{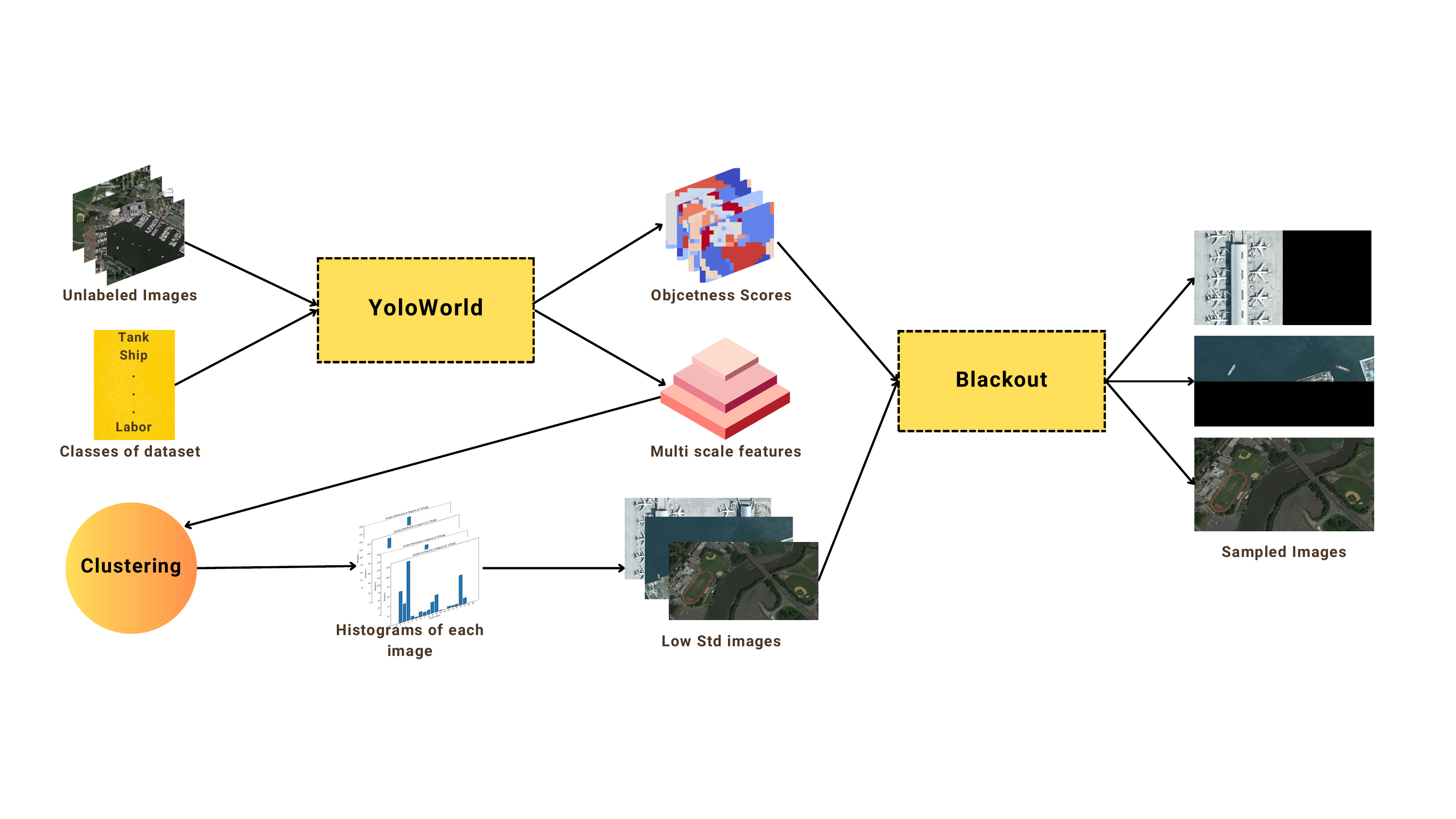} 
\end{graphicalabstract}

\begin{highlights}
\item Reduced labeling effort via robust initial data selection and image refinement.
\item Foundation model integration ensures diverse and high-confidence data selection.
\item Diverse initial data selection enhances model performance in subsequent cycles too.
\end{highlights}

\begin{keyword}
Active learning \sep Object detection \sep Feature-based clustering \sep YOLO-World \sep Objectness scores \sep Consistency-based sampling \sep 


\end{keyword}

\end{frontmatter}



\section{Introduction}
\label{sec1}
The effectiveness of object detection methods mainly depends on training data, and the manual labeling process is expensive, especially in complex environments~\cite{ref1}. As a result, only parts of the data are typically chosen for labeling. Active learning, a strategic approach that selectively annotates the most informative samples, helps with resource limitation problems and improves model accuracy~\cite{ref2,ref3,ref4,ref5}. Active learning performance could be improved by carefully selecting the initial
training samples~\cite{ref6}.
Although some research has been conducted on initial data sampling strategies in natural language processing (NLP) and image classification, the first active learning cycle in object detection has seen relatively less exploration. For example, models such as BERT~\cite{ref7} are used to extract features and select samples that most affect the model for the NLP task~\cite{ref8}. In image classification, strategies such as the K center algorithm have been used for initialization~\cite{ref9}, and covering strategies are utilized to select samples that maximize the coverage of the data distribution~\cite{ref10}. Our framework seeks to fill this research gap in object detection by adapting these proven principles.
Unlike traditional approaches that often rely on random sampling~\cite{ref11,ref12} for initial sampling cycles, our method combines uncertainty sampling with feature-based clustering from the YOLO-World~\cite{ref13} model. The Yolo-world model was chosen for its demonstrated superiority in speed and accuracy compared to other foundation models, especially in zero-shot object detection scenarios. This dual strategy targets highly uncertain samples and ensures diverse representation by using deep feature insights from YOLO-World. Such an approach is specifically practical in early training cycles, ensuring the model can see various object types and scenarios.
The Aquarium~\cite{ref14} and NWPU VHR-10~\cite{ref15} datasets were chosen for their complexity and challenges, such as varied object scales and diverse environmental conditions. These elements make these datasets ideal for demonstrating the robustness and necessity of our approach, especially in managing the inherent difficulties presented by diverse and detailed visual data, even when zero-shot accuracy is lacking. Our method’s ability to tackle these complexities shows its potential to improve how initial training data is selected for active object detection. Here are our contributions to this research:

\begin{itemize}
    \item \textbf{Advanced Sampling Based on YOLO-World Features:} Our approach selects the most informative and diverse training samples by utilizing deep feature representations and clustering algorithms, setting a robust foundation for the model's learning process.
    \item \textbf{Refinement Using YOLO Objectness Scores:} Our method identifies and removes the least confident regions from selected images using objectness scores from YOLO~\cite{ref16}. This ensures the model focuses on areas with higher predictive confidence and helps improve the model's learning process. Concentrating on more reliable regions ensures the training data is relevant and robust, significantly enhancing model accuracy in current and subsequent cycles.
    \item \textbf{Empirical Validation on Complex Datasets:} The effectiveness of our methodology is rigorously tested against the challenging Aquarium and NWPU VHR-10 datasets from different angles, including image-based, object-based, and labeling time-based evaluations. Our findings show significant improvements in training efficiency and model performance, confirming that our method outperforms random sampling.
\end{itemize}

Through these contributions, our proposed framework demonstrates its potential to enhance the early training cycles of object detection models, making it a valuable tool for practitioners dealing with challenging visual data.

\section{Related Works}
\label{sec2}

Active learning (AL) techniques are widely applied in computer vision tasks, such as object detection, to reduce data labeling costs~\cite{ref17, ref18}. The main methods in AL are uncertainty-based sampling, diversity-based sampling, and expected model change. Each addresses different aspects of selecting the most informative data points for model improvement, especially in complex domains like object detection~\cite{ref19}.

\subsection{Uncertainty-Based Sampling}
Uncertainty sampling is one of the most famous AL methods due to its ability to enhance model performance by focusing on ambiguous samples. In this approach, the model identifies samples with the lowest confidence or highest entropy, directing labeling resources toward data points where model predictions are less precise~\cite{ref20}. For object detection tasks, where each image may contain multiple objects with varying degrees of importance, uncertainty sampling must extend beyond single-label predictions to multi-region confidence assessments~\cite{ref21}. Techniques such as SoftMax-based uncertainty and entropy calculations are widely utilized to measure this uncertainty in classification and detection contexts~\cite{ref17}. An advanced form of uncertainty-based sampling is inconsistency in model predictions~\cite{ref21, ref22}. This approach focuses on samples where the model’s predictions are unstable across augmented versions of the same input. For example, in ~\cite{ref23}, bounding box inconsistency is combined with uncertainty metrics to calculate object localization scores within images. However, uncertainty sampling can face limitations when working with small initial labeled pools, as the model’s confidence estimates may be unreliable without enough knowledge~\cite{ref24}.

\subsection{Diversity-Based Sampling}
Diversity-based sampling aims to maximize the variety of labeled data, allowing the model to learn from a broader range of examples~\cite{ref25}. This approach typically uses clustering~\cite{ref26} techniques or distance metrics~\cite{ref27, ref28}  to select distinct samples, which helps to reduce redundancy in the training data. The effectiveness of this method is particularly noticeable during the early training cycles, as exposure to diverse examples significantly enhances the model's robustness~\cite{ref29}.

\subsection{Expected Model Change}
Expected model change, another well-known AL method, emphasizes selecting samples predicted to cause the most significant updates in model parameters~\cite{ref30}. Although effective in targeting influential data, expected model change methods are computationally intensive, requiring estimating parameter changes or loss computations for each candidate sample~\cite{ref31}.

\subsection{Warm and Cold Start Techniques}
A key consideration in AL is the challenge of data scarcity at the initial stages of training that is often addressed through warm and cold start approaches~\cite{ref32}. Warm-start techniques~\cite{ref33, ref34, ref35} rely on a small labeled subset to initiate the training process, which enables the model to produce initial uncertainty measures for further data selection. Cold-start techniques~\cite{ref36, ref37}, by contrast, utilize unsupervised or weakly supervised data to bootstrap the AL process without labeled data~\cite{ref32}. While warm-start methods are effective in AL cycles, they can introduce biases if certain classes are over-represented in the initially labeled subset, particularly in imbalanced datasets~\cite{ref38}.

\subsection{Foundation Models in Active Learning}
Recent studies highlight the integration of foundation models~\cite{ref39, ref40, ref13} within AL frameworks. These models, pre-trained on extensive datasets, offer robust feature representations that facilitate generalization, even when task-specific labeled data is limited. Foundation models in AL settings improve sample selection by leveraging their knowledge to identify complex, high-value samples for labeling. However, the full potential of foundation models has not
been explored in the context of AL, specifically in the low-budget regime~\cite{ref41}.

\section{Methods}
\label{sec3}

\subsection{Overview}
This study introduces how to improve the first active learning cycle using foundational models to select the data for labeling. By getting the initial cycle right, we can improve the early training and set the stage for better results in the later cycles. Foundational models are good at producing high-dimensional representations that capture complex features across various data types. Using these features, we aim to select initial data in active learning so that only the most informative samples are labeled in the first cycle.

\begin{figure*}[!ht]
    \centering
    \includegraphics[scale=0.3]{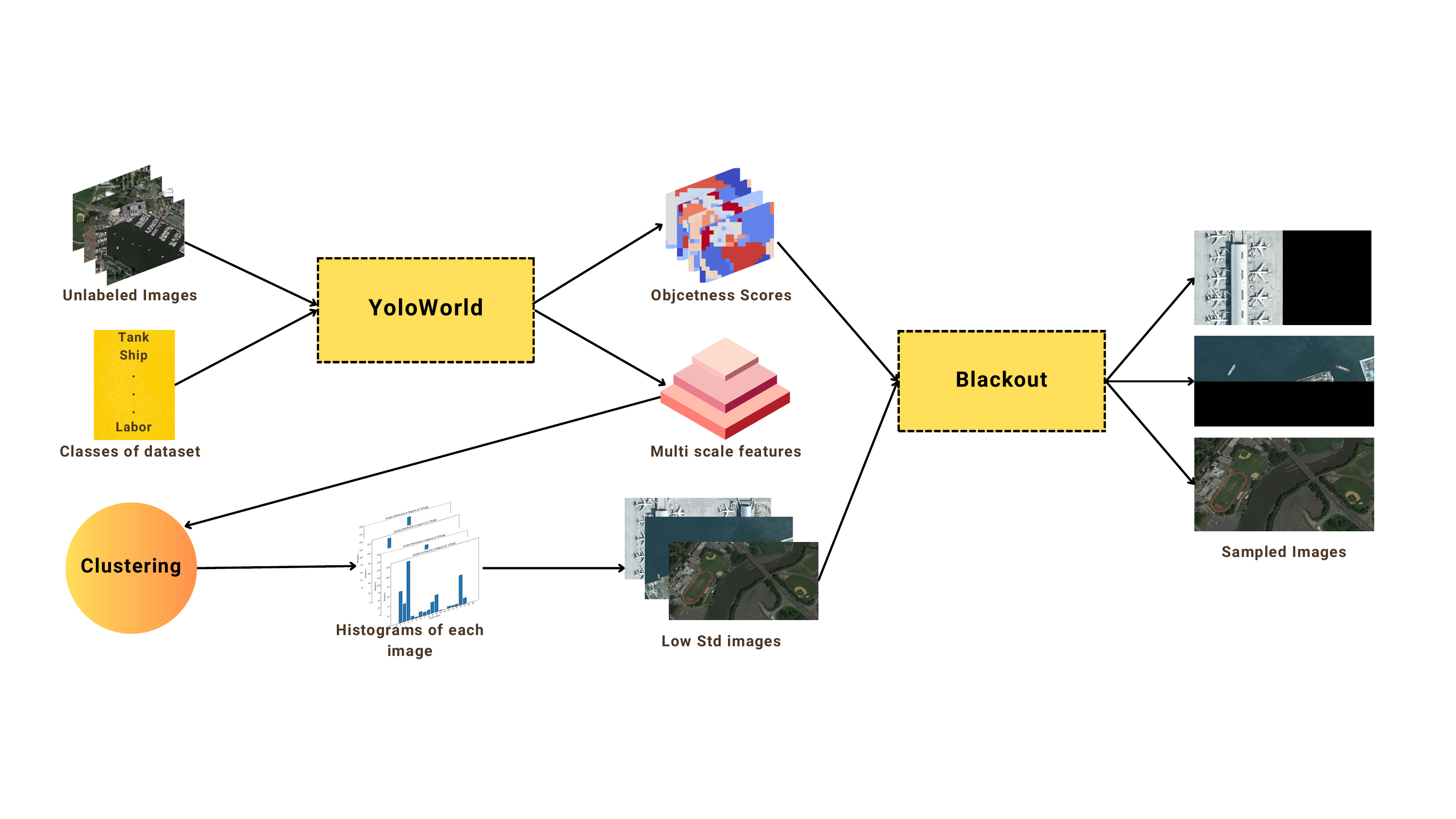} 
    \caption{An overview of the proposed methodology, highlighting the main stages of feature extraction and region selection.}
    \label{fig:methodology_overview}
\end{figure*}

Our approach is a four-stage procedure: \textbf{feature extraction}, \textbf{diversity-based clustering}, \textbf{confidence-based refinement}, and \textbf{iterative evaluation}. Each stage is designed to maximize learning outcomes while minimizing labeling costs. Figure~\ref{fig:methodology_overview} shows an overview of the proposed methodology's main stages of feature extraction and region selection.

\subsection{Feature Extraction Using Foundational Models}
The initial phase of our method employs the YOLOv8~\cite{ref16} vision backbone of the YOLO-World model to extract detailed, multi-scale features from regions of image data. We incorporate class names from our datasets as text prompts to enrich these features. This use of textual information makes the YOLO-World model target class-specific features. The PAN (Pyramid Attention Network) module within YOLO-World further integrates these features, producing enhanced features suitable for our task. These robust features form the foundation for subsequent
clustering. A straightforward clustering algorithm (K-means) organizes these feature vectors into distinct clusters representing varied region types and visual patterns within the dataset. After feature extraction and clustering, each region in the images is associated with one of these clusters.

\subsection{Frequency-Based Sampling}
The frequency-based sampling step of this study aims to choose images that provide the most information for training the model. This selection relies on the distribution of clusters assigned to regions within each image to find the most important and diverse images for the initial active learning cycle. The detailed steps of this process are as follows:

\paragraph{Cluster Frequency Calculation:}
We calculate a frequency vector representing the distribution of clusters assigned to all regions of each image. The frequency vector is defined as:
\begin{equation}
\text{freq}(i) = [f_1, f_2, \ldots, f_k]
\end{equation}
where \( f_k \) denotes the number of regions in image \( i \) that belong to cluster \( k \). This step forms the basis for evaluating each image's diversity of feature representations.

\paragraph{Standard Deviation of Frequencies:}
To assess the variability of cluster distribution within each image, we compute the standard deviation of the frequencies. The standard deviation serves as an indicator of the uniformity of distribution across clusters within an image, with lower values suggesting a more uniform distribution equal to an ideal diverse dataset:
\begin{equation}
\text{std\_devs}(i) = \text{std}(\text{freq}(i))
\end{equation}

\paragraph{Probability Calculation for Image Selection:}
Probability Calculation for Image Selection: We apply a softmax function to the negative standard deviations:
\begin{equation}
    \text{image\_scores} = \text{softmax}(-\text{std\_devs})
\end{equation}
We use negative values of standard deviations and the softmax function to focus on images with more uniform feature distributions. This method focuses on images with a wider range of features, which are likely to have different objects and provide more information for our model. This converts the standard deviations into probabilities, where images with smaller standard deviation values (indicating higher diversity) receive higher probabilities.

\subsection{Enhanced Sampling Strategy for Initial Cycle}

To improve the selection of images in the first cycle of active learning, we incorporate a strategy that considers both the uniformity of feature distribution and the diversity of previously labeled images. Our sampling process follows these steps:

\begin{enumerate}
    \item The process begins by selecting the image with the highest score derived from $\text{std\_devs\_probabilities}$.
    \item Next, the cluster frequencies of each unlabeled image are combined with the frequencies of already labeled images:
    \begin{equation}
        \text{freq}(i) = [f_1,f_2,\dots,f_K ] + [f_1^{'},f_2^{'},\dots,f_K^{'}]
    \end{equation}
    where $f_k$ represents the frequency of occurrences of cluster $k$ in the unlabeled image $i$, and $f_k^{'}$ denotes the frequency of cluster $k$ in the previously labeled dataset. The summation provides an updated frequency distribution that integrates both labeled and unlabeled image features.
    \item Using this updated frequency, a new standard deviation is computed:
    \begin{equation}
        \text{std\_devs}(i) = \text{std} (\text{freq}(i))
    \end{equation}
    \item A SoftMax function is applied to the negative standard deviation values to generate scores that reflect how well each image complements the labeled pool:
    \begin{equation}
        \text{pool\_score} = \text{SoftMax}(-\text{std\_devs})
    \end{equation}
    \item Finally, a total score for each image is computed by combining its initial uniformity score and the new diversity score:
    \begin{equation}
        \text{Total\_scores} = 0.5 \times \text{std\_probs} + 0.5 \times \text{new\_scores}
    \end{equation}
\end{enumerate}

This iterative process continues until the labeling budget is met. Higher scores indicate that selecting the new image will lead to a more uniform distribution of labeled image features. After selecting the highest-scoring image, the process repeats for the next selection, ensuring diversity and balanced representation in the labeled dataset.

By integrating this method, we optimize the first cycle of active learning, ensuring that the model benefits from a well-distributed and informative set of labeled images, leading to improved performance in subsequent learning cycles.

\subsection{Image Refinement Based on Objectness Scores}
In the final stage of our method, we refine the selected images by targeting regions with the lowest objectness scores from the YOLOv8 vision backbone. These regions, characterized by low confidence levels in the model, indicate that our foundational model has struggled to extract meaningful features, making it unlikely that these areas contain relevant objects or that the objects present provide sufficient information for adequate labeling. After removing these low-confidence areas, the model can concentrate on more informative regions in subsequent training cycles. This approach helps reduce redundancy in object types across the dataset, preventing the model from overfitting on frequently appearing objects. It also eliminates large, empty, or irrelevant spaces that do not contribute to effective training. Additionally, this can be considered a form of data augmentation. By refining the selection of image regions, our method ensures that the labeling budget is spent on more diverse and crucial objects.
 
\paragraph{Objectness Score Calculation:}
The objectness score for each quadrant of an image is calculated by averaging YOLOv8's objectness scores across all predictions in that quadrant as:
\begin{equation}
C_{\text{quadrant}} = \frac{1}{n_{\text{quadrant}}} \sum_{i=1}^{n_{\text{quadrant}}} \text{objectness}_i
\end{equation}
Here, \( n_{\text{quadrant}} \) denotes the number of predictions in the specific quadrant and \( objectness_{\text{i}} \) is the objectness score for the \( i \) prediction.

\paragraph{Average Objectness Score Across the Image:}
Following this, an average objectness score is computed for the entire image to set a benchmark for comparison: 
\begin{equation}
C_{\text{avg}} = \frac{1}{n_{\text{total}}} \sum_{i=1}^{n_{\text{total}}} \text{conf}_i
\end{equation}
Where \( n_{\text{total}} \) is the total number of predictions across the image.

\paragraph{Comparative Elimination:}
The refinement strategy involves comparing the average objectness score of each quadrant against the image's overall average objectness score. Quadrants with an average score lower than the image's average are identified for potential elimination. This decision criterion ensures that regions within an image deemed less likely to contain objects are selectively removed.

\begin{figure}[h!]
    \centering
    \includegraphics[scale=0.5]{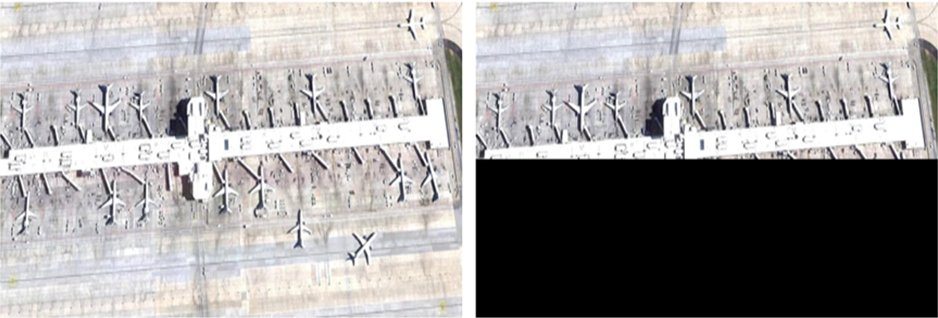} 
    \includegraphics[scale=0.5]{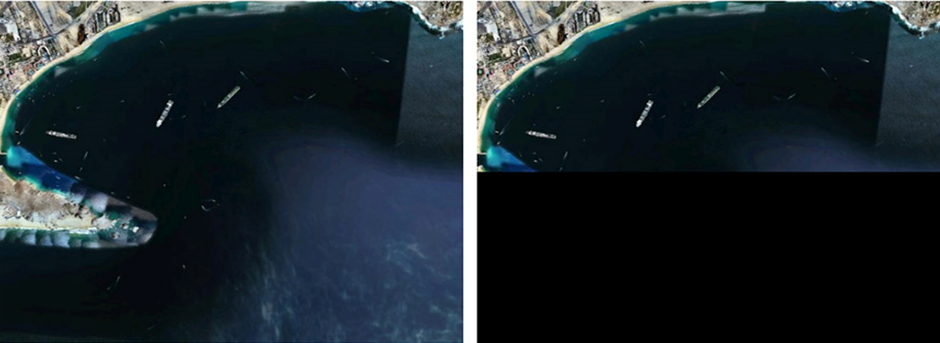} 
    \caption{Detailed visualization showing refined regions after applying objectness score-based elimination.}
    \label{fig:refinement}
\end{figure}

\paragraph{Selective Refinement:}
This targeted refinement strategy leads to a more focused training dataset where only regions with higher objectness scores are retained. By doing so, the model dedicates more computational resources to areas more likely to improve its detection capabilities in subsequent iterations. After regions with lower scores are removed, the corresponding labels are updated to reflect the refined dataset, ensuring that subsequent training phases concentrate on improving predictions in higher-confidence areas. To demonstrate this refinement process, selected images showing regions that have been removed are shown in Figure~\ref{fig:refinement}. These visualizations show how objectness scores make refinement on initial sampled data.

\section{Experiments}
In this section, we evaluate the performance of our proposed method for active object detection by comparing it with other image sampling techniques, specifically focusing on the initial sampling cycle. We utilize The NWPU VHR-10~\cite{ref42, ref15} and the Aquarium dataset~\cite{ref14}. Each dataset is selected to assess the method's efficacy under varied and complex conditions, challenging the detection models with diverse environmental factors.\newline
\textbf{Datasets}: The NWPU VHR-10 dataset, a benchmark in remote sensing and satellite image processing, comprises 800 high-resolution aerial images distributed across ten classes, including aeroplanes, ships, and vehicles. Satellites and drones collect These images under various lighting and angle conditions. They are particularly challenging due to the small size or poor illumination of the objects depicted. We have selected 200 images from this dataset to use as test data. The Aquarium dataset, provided and labeled by Roboflow, includes 638 images of marine life from two U.S. aquariums, containing seven classes, such as fish, jellyfish, and penguins. One hundred ninety-one images are used as test data; the rest is considered training data, and in our work, we use it as unlabeled data.\newline
\textbf{Settings}: For our experiments, the number of clusters is determined to be twice the number of classes in each dataset, ensuring a comprehensive coverage of the data characteristics. We employ YOLOv8 as our primary detection model, configured with a batch size of 16 and a learning rate of 0.01. Our sampling methodology for the first cycle of active learning employs our proposed technique, while for subsequent cycles, we use a method used in \cite{ref21} that utilizes the consistency of the trained model to sample images for the next cycle. We report the average mAP50 with standard deviation, conducting at least four trials for each experiment to ensure reliability and reproducibility. Additional details about the experimental settings and configurations are discussed in the respective sections dedicated to those experiments.

\begin{table}[t]
\centering
\footnotesize
\begin{tabular}{|c|c|c|c|c|c|}
\hline
\multirow{2}{*}{Method} & \multicolumn{5}{c|}{Map50 in \% (NWPU dataset)} \\
\cline{2-6}
& cycle0 & cycle1 & cycle2 & cycle3 & cycle4 \\
\hline
Random & 37.2±4.8 & 37.5±3.6 & 41.6±3.8 & 48.9±12.1 & 56.4±10.7 \\
Clustering (MaxBin) & 42±3.6 & \textbf{44.5±3.4} & \textbf{50.7±4} & 52.8±1.6 & 54.7±2.6 \\
Clustering (STD) & \textbf{43.8±3.6} & \textbf{44.5±2.7} & 49.5±5.9 & \textbf{53.3±4.5} & \textbf{56.9±5.9} \\
Clustering + blackout & 29.1±3.4 & 33.6±4.5 & 44.2±5.2 & 46.7±2.9 & 52.6±6.3 \\
\hline
\end{tabular}
\caption{Performance comparison (Map50 in \%) for NWPU dataset.}
\label{table:nwpu_sampling_comparison}
\end{table}

\begin{table}[t]
\centering
\footnotesize
\begin{tabular}{|c|c|c|c|c|c|}
\hline
\multirow{2}{*}{Method} & \multicolumn{5}{c|}{Map50 in \% (Aquarium dataset)} \\
\cline{2-6}
& cycle0 & cycle1 & cycle2 & cycle3 & cycle4 \\
\hline
Random & 25.2±3.5 & 27.2±2.4 & 30.7±5.6 & 32.8±5.1 & 39.6±1.5 \\
Clustering (MaxBin) & \textbf{30.9±3.6} & \textbf{31.6±5.5} & \textbf{35.1±3.3} & 35.6±2.3 & \textbf{40.9±2.3} \\
Clustering (STD) & 29.1±1.2 & 29.1±1.2 & 34.9±3.2 & \textbf{36.0±4.5} & 40.8±2.2 \\
Clustering + blackout & 15.9±1.9 & 20.6±2.3 & 19.6±2.4 & 32.9±3.7 & 37.2±4.2 \\
\hline
\end{tabular}
\caption{Performance comparison (Map50 in \%) for Aquarium dataset.}
\label{table:aquarium_sampling_comparison}
\end{table}
\subsection{Image Selection Performance Comparison}
\begin{figure}[h!]
    \centering 
    \begin{subfigure}[t]{\textwidth} 
        \centering
        \includegraphics[height=6cm, width=10cm]{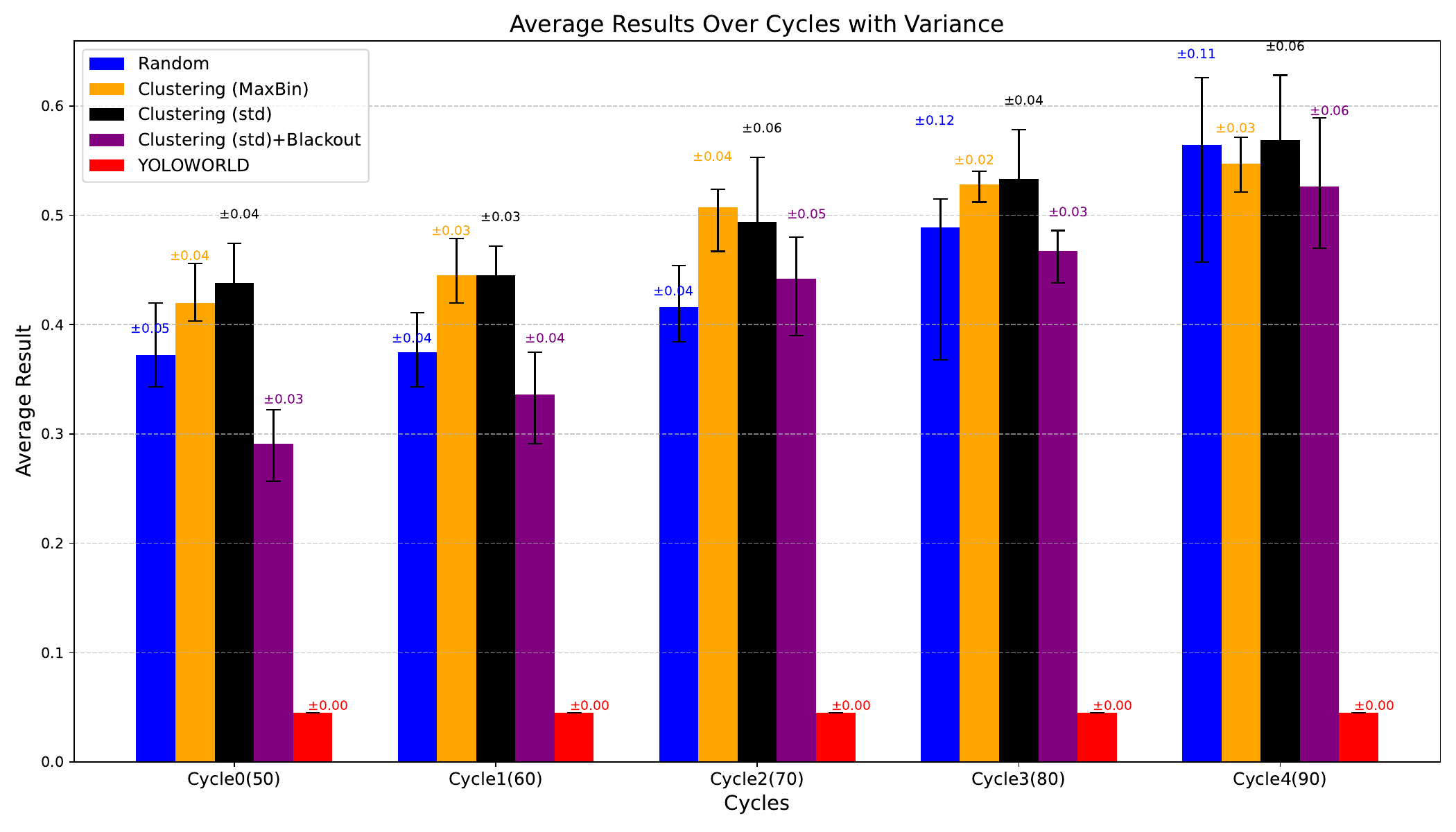} 
        \caption{Results on Aquarium dataset}
    \end{subfigure}
    \vspace{10pt} 
    \begin{subfigure}[t]{\textwidth} 
        \centering
        \includegraphics[height=6cm, width=10cm]{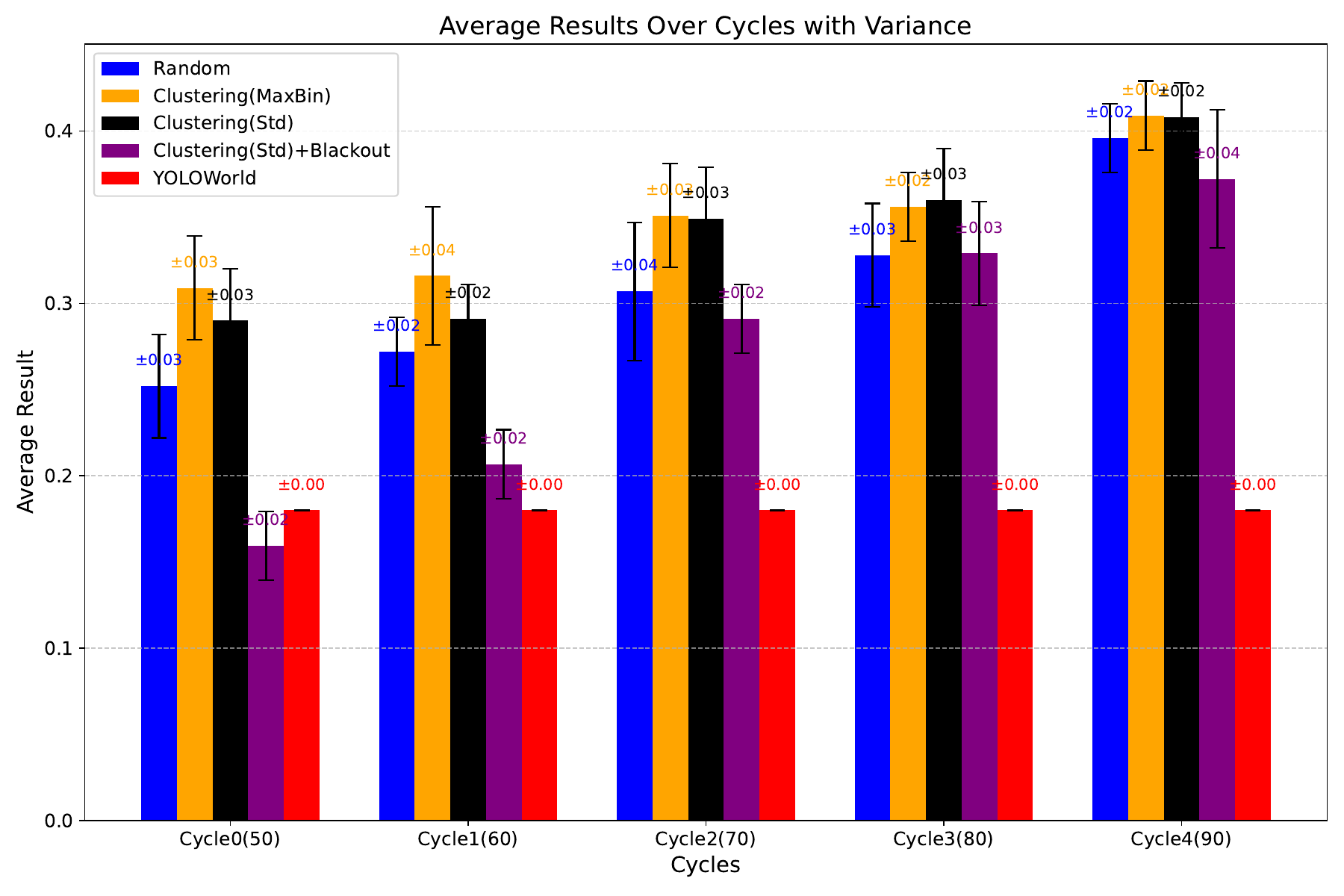} 
        \caption{Results on NWPU dataset}
    \end{subfigure}
    \caption{Comparison of our method to different sampling methods based on number of images.}
    \label{fig:clustering_refinement}
\end{figure}

In this subsection, we compare the effectiveness of our proposed Standard Deviation (STD) method against other sampling strategies in the initial cycle of image selection. We contrast it with random sampling(random for the first cycle and uncertainty sampling~\cite{ref21} for other cycles) and the MaxBin method, which selects based on the highest number of clusters formed after clustering. Our STD approach distinctively selects images with the highest variance in cluster frequency. This technique captures more informative samples by focusing on cluster variability rather than sheer quantity.
Additionally, the Clustering (STD)+blackout variant refines this approach by excluding regions within the images with low confidence levels, thus concentrating the learning process on areas where the model shows more uncertainty. We increase the number of selected samples each cycle, beginning with 50 images in the first cycle and incrementing by 10 in each cycle. The model is trained for 20 iterations for each cycle, and the average results of five trials for each method are reported. Table~\ref{table:nwpu_sampling_comparison} and table~\ref{table:aquarium_sampling_comparison} display the results of these experiments, reporting the mean accuracy and standard deviations for various sampling strategies. Figure~\ref{fig:clustering_refinement}, for each dataset, shows the improvement in model accuracy over the cycles for each sampling method.

\begin{figure}[h!]
    \centering
    \begin{subfigure}[t]{0.8\textwidth} 
        \centering
        \includegraphics[width=\textwidth]{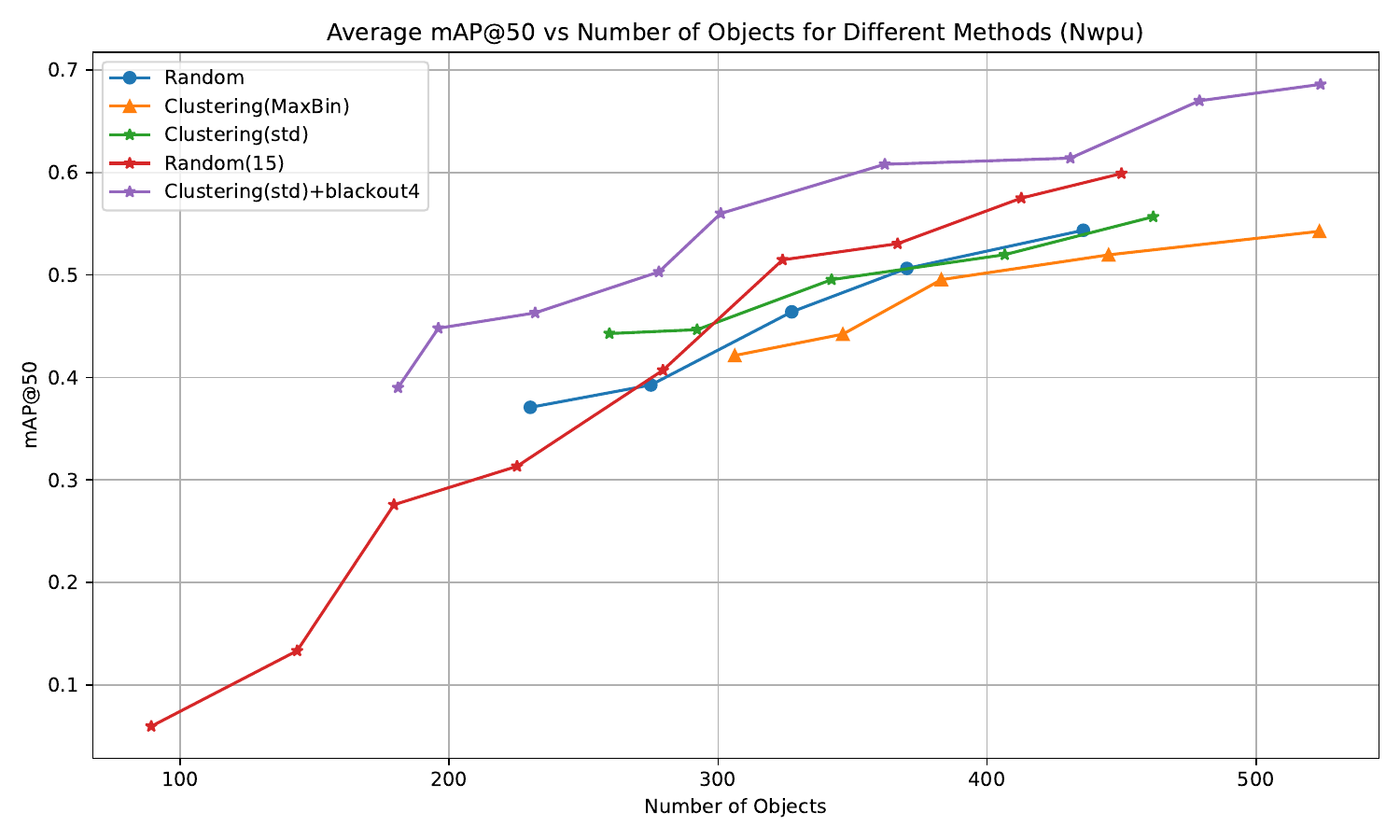} 
        \caption{}
    \end{subfigure}
    \vspace{10pt} 
    \begin{subfigure}[t]{0.8\textwidth} 
        \centering
        \includegraphics[width=\textwidth]{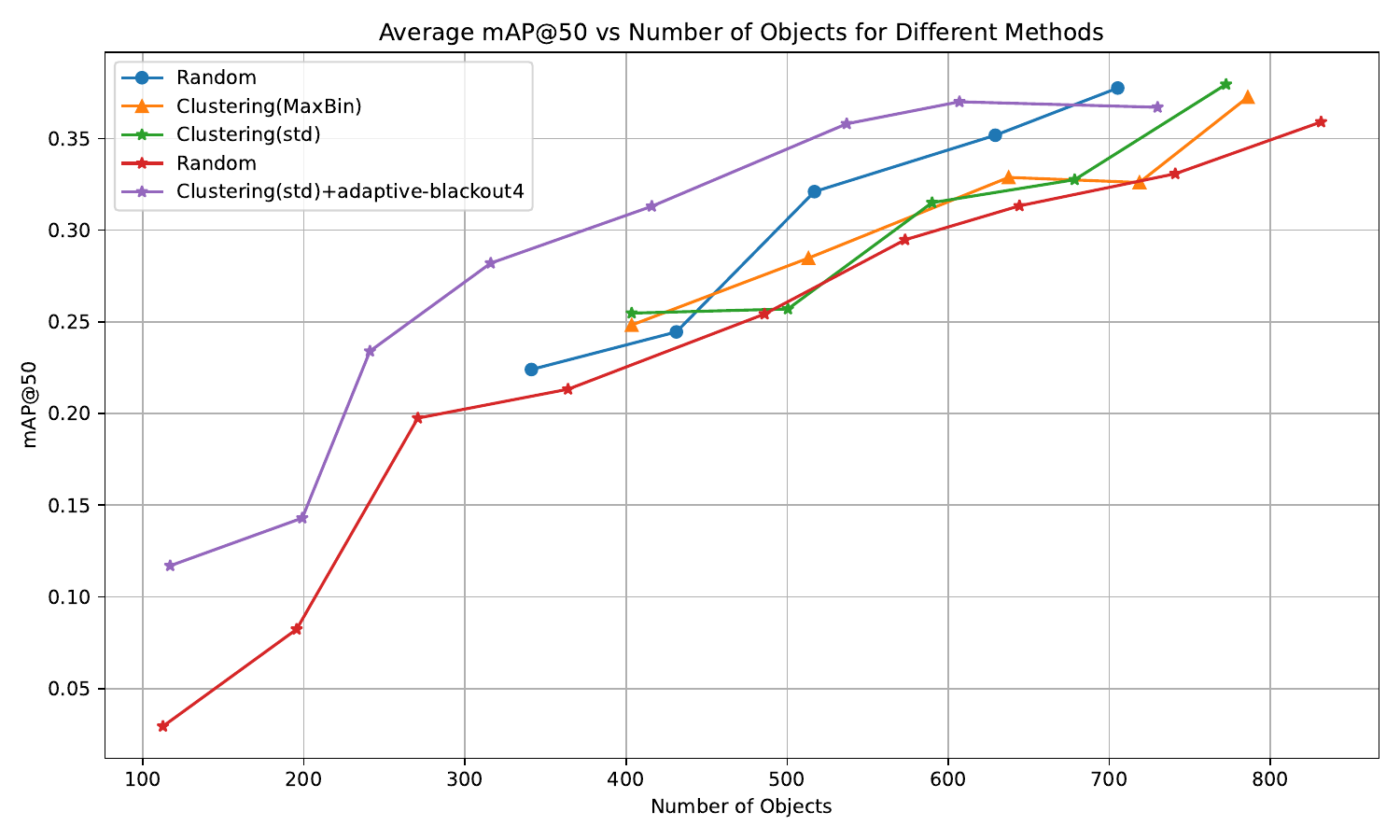} 
        \caption{}
    \end{subfigure}
    \caption{Comparison of our method to different sampling methods based on the number of objects. a) Results on NWPU dataset b) Results on Aquarium dataset}
    \label{fig:objectbased_comparison}
\end{figure}
\subsection{Object Count-Based Performance Analysis}
In this experiment, we evaluate our methods based on the number of objects instead of the number of images. This evaluation method is essential for assessing the actual effectiveness of the object detection system. As noted in \cite{ref12}, comparing methods at the object level is essential to truly understand the practical implications of different sampling strategies on detection performance. In this experiment, we continue with the Standard Deviation(STD) method, MaxBin, and Clustering(STD)+blackout and introduce a new variant, "Random (15)." This method starts with 15 random images in the initial cycle, compared to 50 in the standard setup, aligning the number of objects more closely with those in the Clustering(STD)+blackout and uses uncertainty sampling~\cite{ref21} for other cycles for a fair comparison. We maintain the use of YOLOv8 as our detection model. The results are the averages of five independent trials for each method to ensure statistical reliability. Figure~\ref{fig:objectbased_comparison} will graphically show trends and performance variances across cycles for each dataset.

\begin{figure}[h!]

    \centering
    \begin{subfigure}[t]{0.8\textwidth} 
        \centering
        \includegraphics[width=\textwidth]{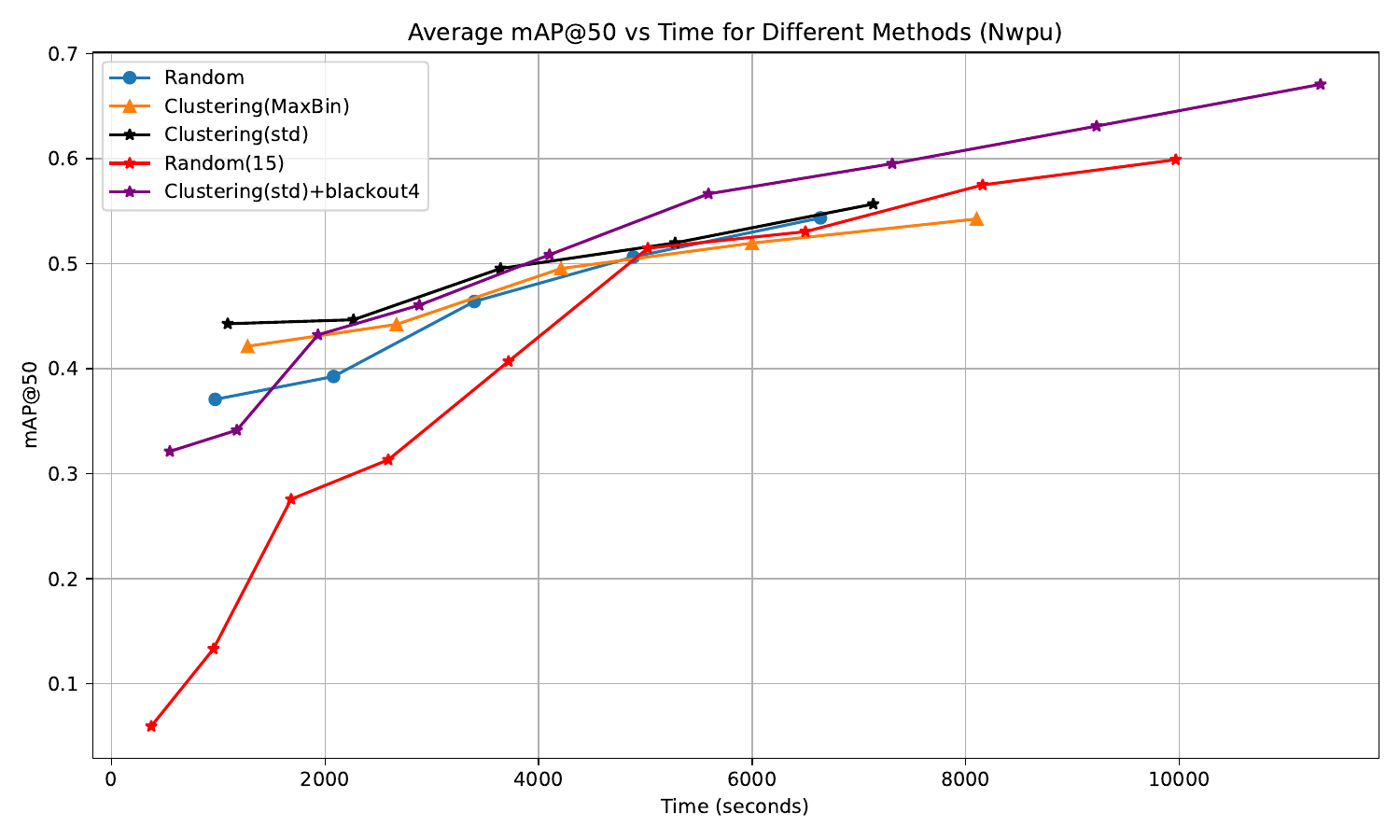} 
        \caption{}
    \end{subfigure}
    \vspace{10pt} 
    \begin{subfigure}[t]{0.8\textwidth} 
        \centering
        \includegraphics[width=\textwidth]{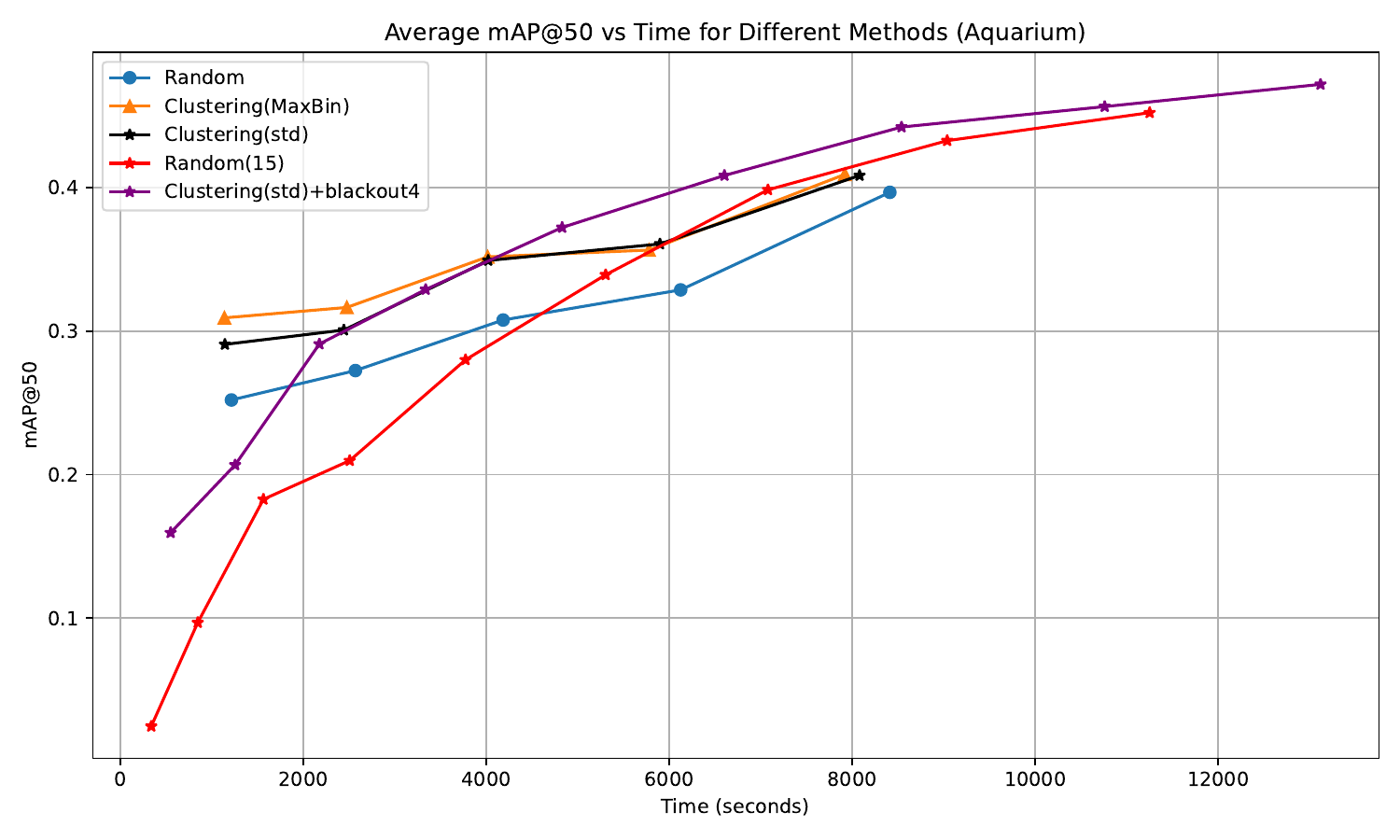} 
        \caption{}
    \end{subfigure}
    \caption{Comparison of our method to different sampling methods based on time spent for labeling. a) Results on NWPU dataset b) Results on Aquarium dataset}
    \label{fig:time_comparison}
\end{figure}

\subsection{Labeling Time Efficiency Analysis}
In this experiment, we evaluate the time efficiency of different sampling methods by analyzing the labeling process for the selected datasets. Each bounding box annotation required an average of four seconds, with an additional second needed to transition between images. The labeling process was performed using the LabelMe \cite{ref43} tool, a widely used annotation utility in machine learning tasks.
To assess efficiency, we measured and compared the total time required to label images selected by various sampling strategies, including Standard Deviation (STD), MaxBin, Clustering (STD) + blackout, and the "Random (15)" variant. In addition to the manual labeling time, the total time accounts for the computational cost of each sampling method. These time efficiency metrics provide insights into which sampling strategy best balances detection performance and operational efficiency. The comparative results are visually represented in Figure~\ref{fig:time_comparison}, illustrating the efficiency of each method across different cycles.

\subsection{Supplementary Experiments}
This section consolidates further experiments designed to evaluate specific aspects of the proposed method and other techniques in depth.

\subsubsection{Cluster-Class Relationship Analysis}
\begin{figure}[h!]
    \centering
    \begin{subfigure}[t]{0.4\textwidth} 
        \centering
        \includegraphics[width=\textwidth]{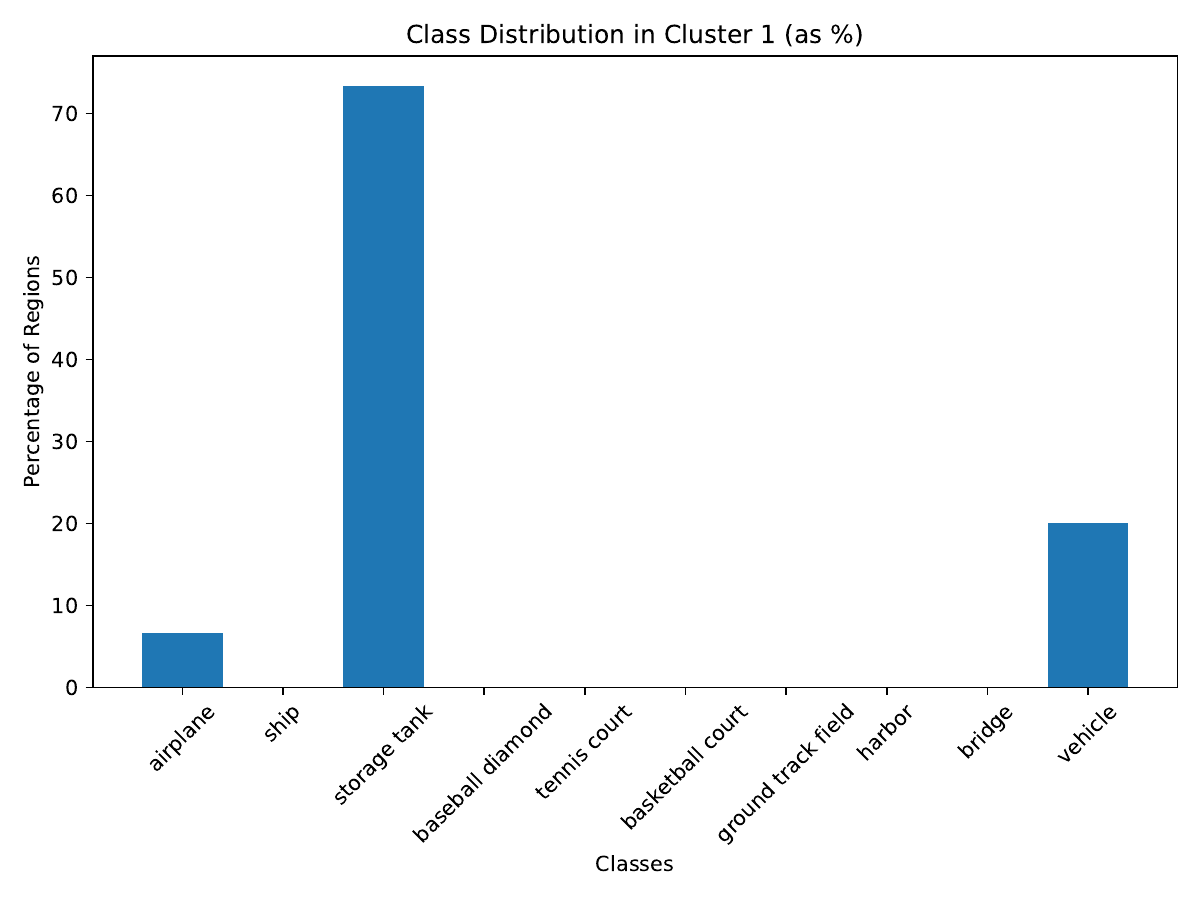} 
        \caption{}
    \end{subfigure}
    \hfill
    \begin{subfigure}[t]{0.4\textwidth} 
        \centering
        \includegraphics[width=\textwidth]{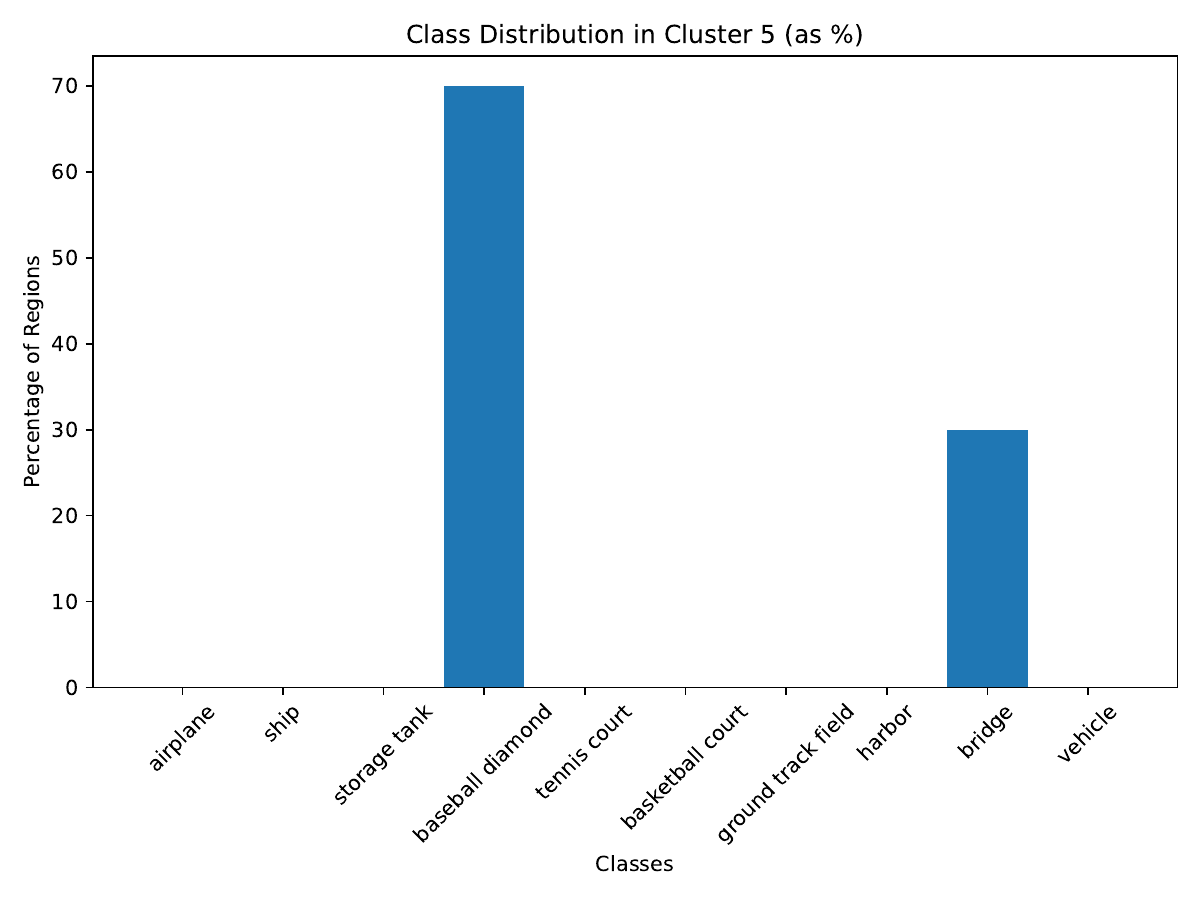} 
        \caption{}
    \end{subfigure}

    \vspace{0.5cm} 
    \begin{subfigure}[t]{0.4\textwidth} 
        \centering
        \includegraphics[width=\textwidth]{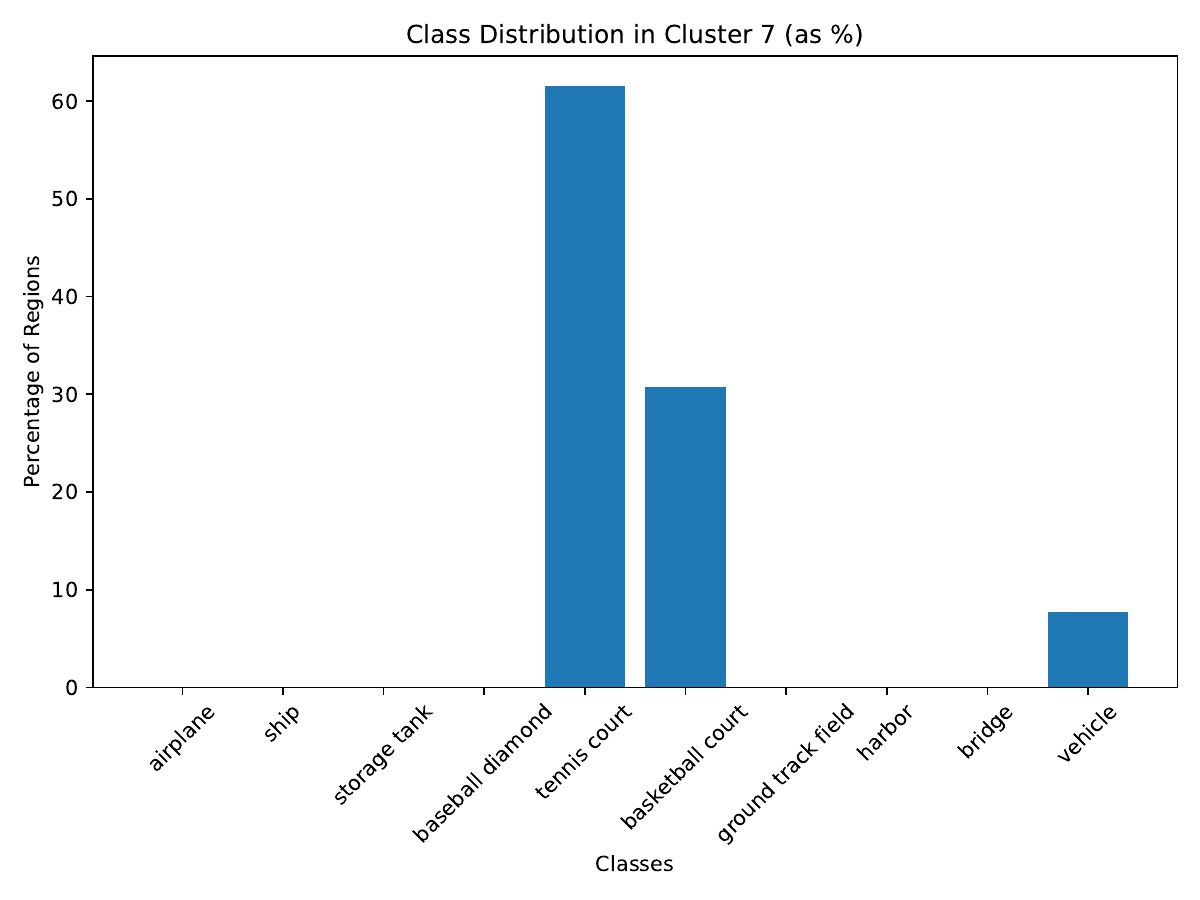} 
        \caption{}
    \end{subfigure}
    \hfill
    \begin{subfigure}[t]{0.4\textwidth} 
        \centering
        \includegraphics[width=\textwidth]{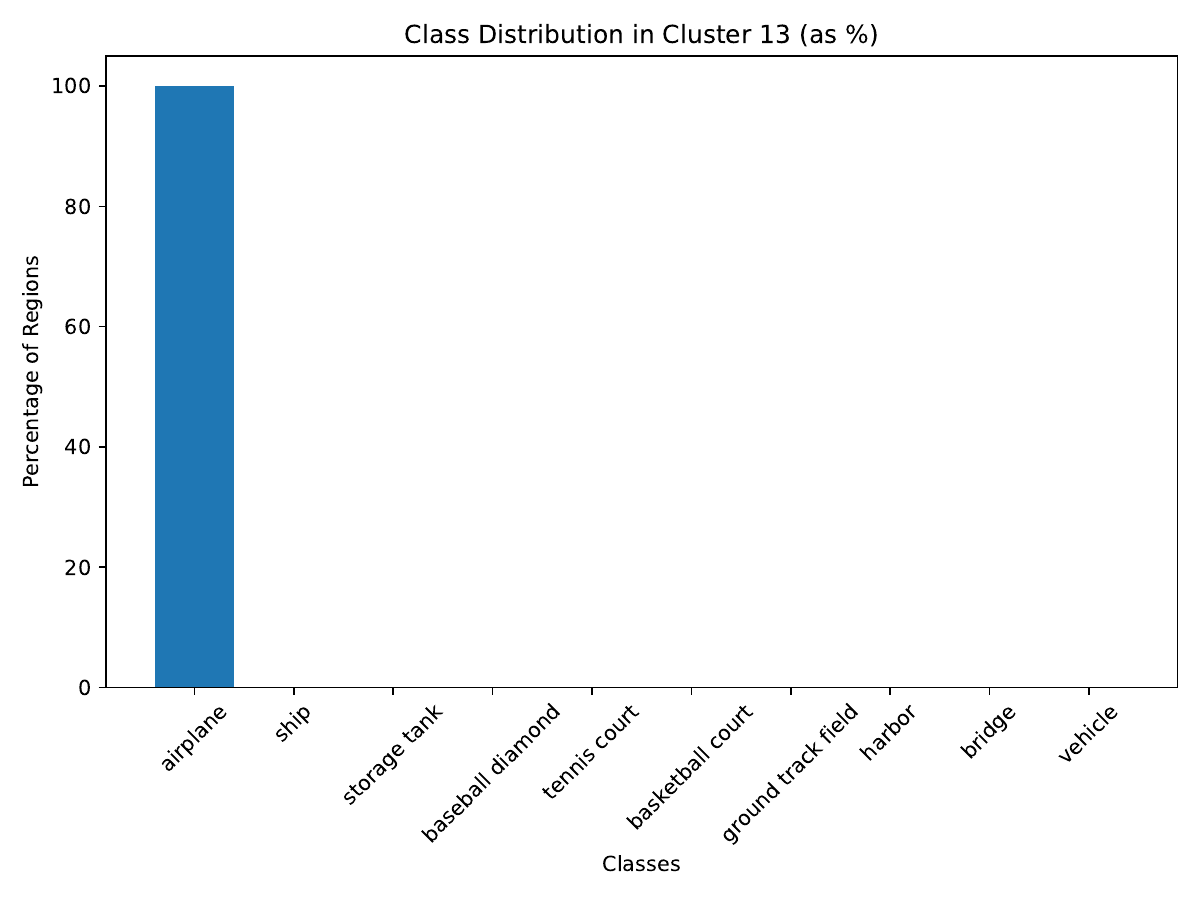} 
        \caption{}
    \end{subfigure}

    \caption{Histograms of each cluster based on classes of regions that contain object.}
    \label{fig:true_label}
\end{figure}

We investigated the relationship between image clusters and the object classes they contain by assigning each region containing an object both a cluster label and a class label. We then examined the distribution of actual object classes within these clusters, visualizing the results with histograms that depict the frequency of each class per cluster, as shown in Figure~\ref{fig:true_label}.
The analysis indicates that our clustering method promotes greater diversity in image selection, with specific clusters predominantly associated with particular object classes. These findings suggest that our approach not only enhances the variety of images chosen but also improves class-specific clustering accuracy, thereby boosting the precision of object detection by aligning clusters more closely with the underlying object features.

\subsubsection{Impact of Removing Certain Image Areas}
\begin{figure}[h!]
    \centering
    \begin{subfigure}[t]{0.9\columnwidth} 
        \centering
        \includegraphics[width=\textwidth]{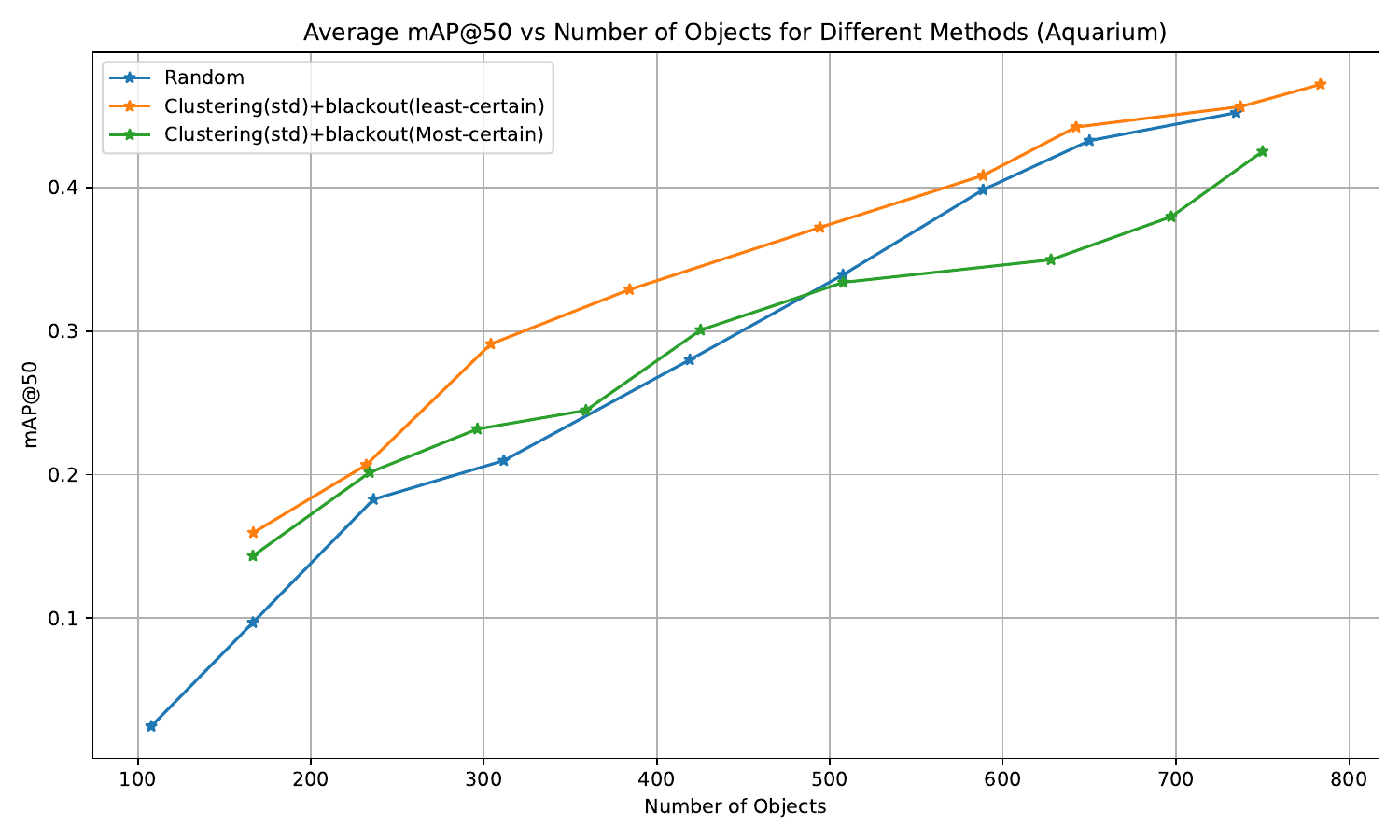} 
        \caption{}
    \end{subfigure}
    \vspace{5pt} 
    \begin{subfigure}[t]{0.9\columnwidth} 
        \centering
        \includegraphics[width=\textwidth]{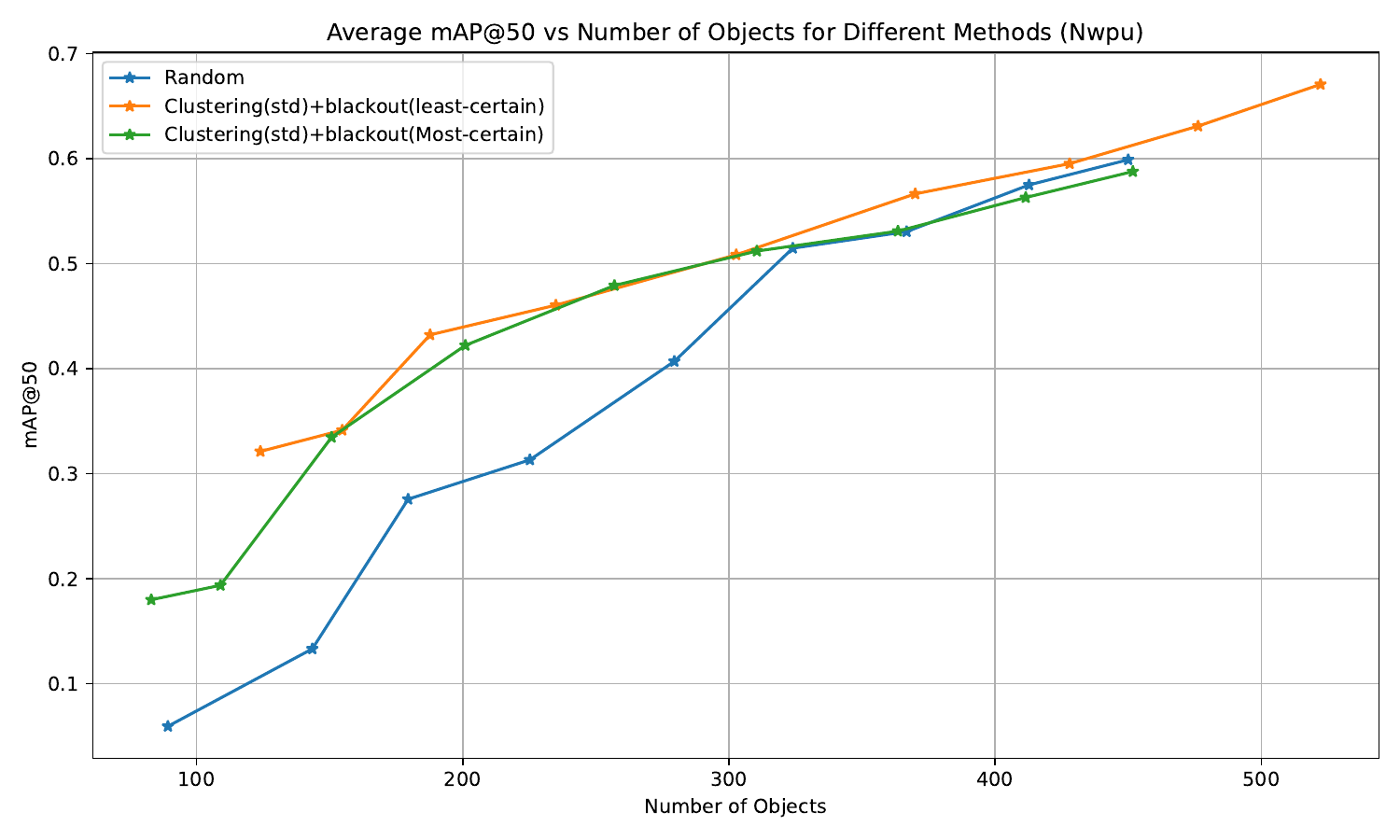} 
        \caption{}
    \end{subfigure}
    \caption{Comparison on results of removing uncertain parts of sampled images (clustering(std)+blackout(least-certain)) with removing certain parts of sampled images (clustering(std)+blackout(most-certain)). a) On Aquarium dataset b) On NWPU dataset.}
    \label{fig:certain_delete}
\end{figure}

This experiment explores the impact of removing high-confidence areas from images instead of the conventional method of excluding low-confidence regions. We identified high-confidence areas based on average confidence scores that indicate the presence of objects and removed these regions from the training data. The goal was to compare the system's performance when high-confidence areas are excluded versus when low-confidence areas are removed. As shown in Figure~\ref{fig:certain_delete}, excluding high-confidence regions resulted in worse performance compared to removing low-confidence areas. This suggests that retaining the less certain regions may benefit the model's learning process by encouraging it to focus on and improve its weaker areas, ultimately enhancing accuracy and robustness.

\begin{figure}[h!]
    \centering
    \begin{subfigure}[t]{0.8\textwidth} 
        \centering
        \includegraphics[width=\textwidth]{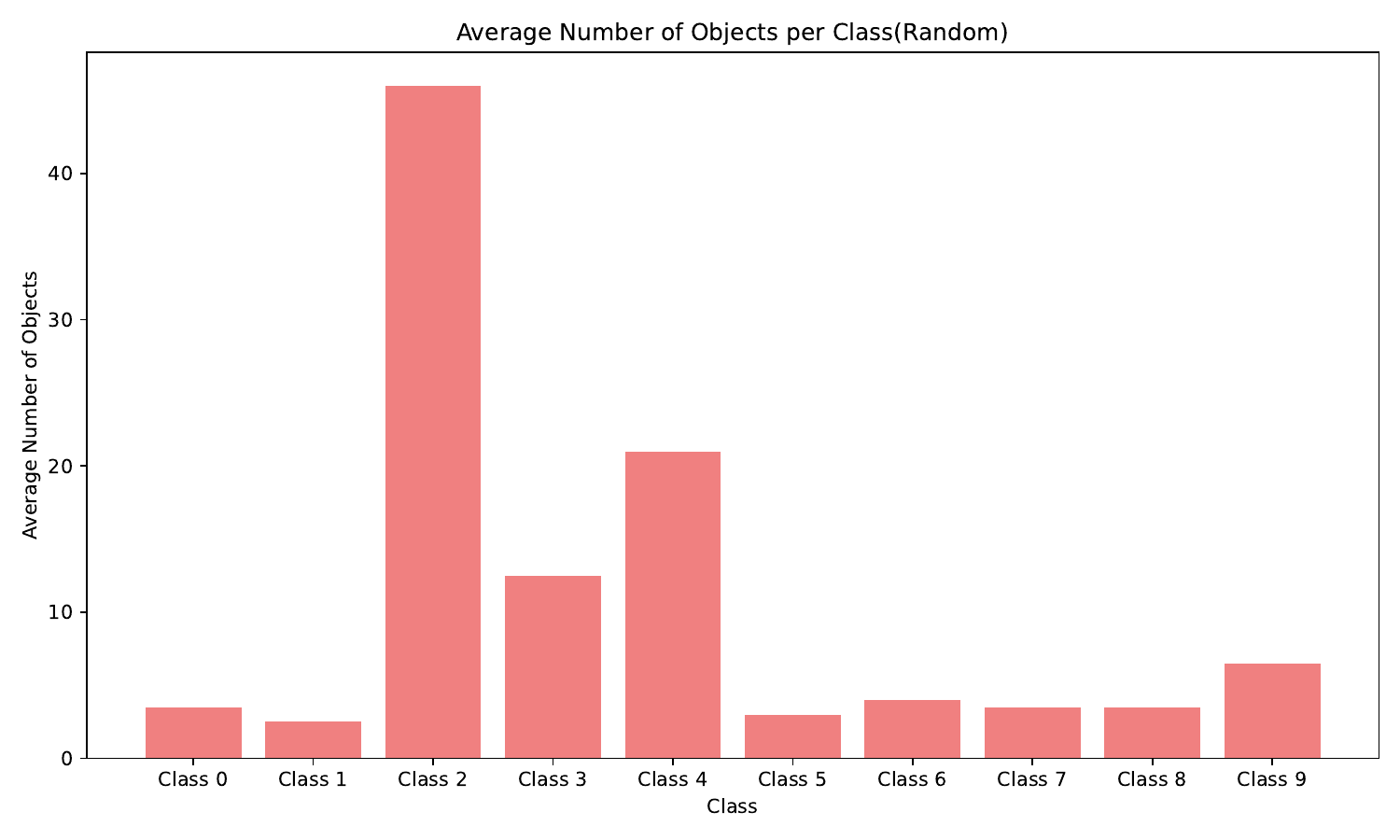} 
        \caption{}
    \end{subfigure}
    \vspace{10pt} 
    \begin{subfigure}[t]{0.8\textwidth} 
        \centering
        \includegraphics[width=\textwidth]{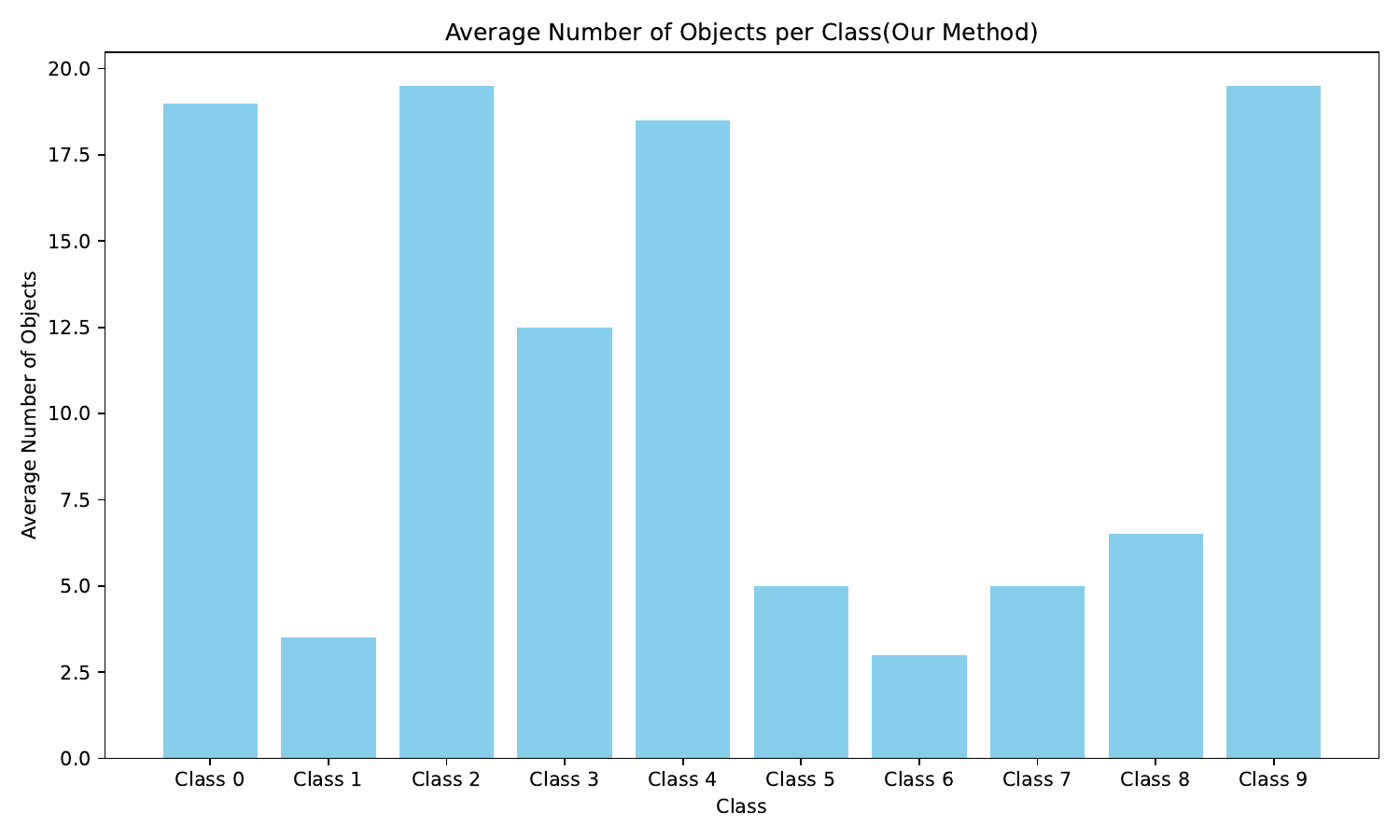} 
        \caption{}
    \end{subfigure}
    \caption{Distribution of sampled data. a) Using random sampling b) Using Our Method (clustering(std)+blackout)}
    \label{fig:distribution}
\end{figure}

\subsubsection{Comparison of the Distribution of Images Selected by the Proposed Method and Random Sampling}
This experiment, conducted across five trials, assesses the efficacy of our proposed sampling method versus random sampling in distributing objects uniformly across classes in the NWPU dataset. The results presented in Figure~\ref{fig:distribution} show that our method yields a more balanced distribution of objects across different classes when compared to random sampling. This balance is necessary during the initial cycle, as it ensures that the model encounters a variety of objects from multiple classes, thereby enhancing its generalization capabilities. Such diversity in early training phases is fundamental for developing robust machine learning models that perform well in varied real-world conditions.

\section{Conclusion}

This study presents a novel active learning framework that optimizes the initial training cycle for object detection models using the Aquarium and NWPU VHR-10 datasets. Our approach leverages feature-based clustering from YOLO-World and refines training images based on objectness scores, enhancing sample selection by prioritizing informative and diverse representations.
We evaluated our method through various experiments, including image-based and object-based analyses and labeling time assessments. Our results demonstrate clear advantages over traditional random sampling in the first training cycle. Furthermore, improvements in subsequent cycles suggest that better selection of the initial data selection plays a crucial role in the effectiveness of active learning.

\end{document}